\documentclass[10pt,twocolumn,letterpaper]{article}

\PassOptionsToPackage{table,dvipsnames}{xcolor}
\usepackage{xcolor}

\usepackage[pagenumbers]{cvpr} % To force page numbers, e.g. for an arXiv version

\definecolor{cvprblue}{rgb}{0.21,0.49,0.74}
\usepackage[pagebackref,breaklinks,colorlinks,allcolors=cvprblue]{hyperref}
\usepackage{makecell}
\usepackage{multirow}
\usepackage{booktabs}
\usepackage{pifont}
\def\paperID{*****} % *** Enter the Paper ID here
\def\confName{CVPR}
\def\confYear{2025}

\title{THU-Warwick Submission for EPIC-KITCHEN Challenge 2025: \\
Semi-Supervised Video Object Segmentation}

\author{Mingqi Gao$^{1,2}$ \quad Haoran Duan$^{1}$ \quad Tianlu Zhang$^{1}$ \quad Jungong Han$^{1,}$\footnotemark[1]\\
$^{1}$Tsinghua University \quad $^{2}$University of Warwick\\
{\tt\small mingqi.gao@warwick.ac.uk, tlzhang@mail.tsinghua.edu.cn, \{haoranduan,jghan\}@tsinghua.edu.cn}
}

\begin{document}
\maketitle
\renewcommand{\thefootnote}{\fnsymbol{footnote}}
% \footnotetext[2]{Equal contributions.}
\footnotetext[1]{Corresponding author.}
\renewcommand{\thefootnote}{\arabic{footnote}}
\begin{abstract}
In this report, we describe our approach to egocentric video object segmentation. Our method combines large-scale visual pretraining from SAM2 with depth-based geometric cues to handle complex scenes and long-term tracking. By integrating these signals in a unified framework, we achieve strong segmentation performance. On the VISOR test set, our method reaches a $\mathcal{J}\&\mathcal{F}$ score of 90.1\%.
\end{abstract}    
\section{Introduction}
\label{sec:intro}

Egocentric visual understanding~\cite{plizzari2024outlook,grauman2024ego,perrett2025hd} enables intelligent systems to perceive the world from a human-centred perspective. It is vital in applications such as embodied AI, augmented reality, and assistive technologies. In these scenarios, most tasks are object-centric, focusing on recognising and tracking the objects the user sees or interacts with. Therefore, achieving pixel-level object perception and consistent tracking over time is crucial. This is the right goal of EPIC-KITCHENS VISOR~\cite{VISOR2022}: to segment and track hands and active objects in egocentric videos. 

Compared to traditional video object segmentation tasks that focus on third-person views~\cite{davis, xu2018youtube, MOSE, hong2024lvos}, VISOR presents unique challenges: 1) The egocentric viewpoint leads to rapidly changing and cluttered backgrounds, making it hard to distinguish target objects and 2) Frequent interactions between the user’s hands and objects cause severe and dynamic occlusions, further complicating accurate segmentation. These challenges require methods that can robustly handle complex scenes with heavy occlusions and ambiguous object boundaries. 

To address these challenges, our solution integrates depth to provide complementary geometric cues that help distinguish target objects from complex backgrounds and handle occlusions more effectively. Specifically, we leverage Depth Anything V2~\cite{depth_anything_v2}, a large-scale depth model, and combine its backbone features with those from a visual backbone. This fusion is performed through a learnable, multi-scale module, enabling the model to utilise depth and RGB features to enhance the segmentation process.

Benefiting from large-scale data, SAM2~\cite{ravi2024sam} has shown strong performance across various video object segmentation tasks. However, their use of segmentation history remains relatively shallow, relying more heavily on recent predictions. This makes it less effective in long-term, complex scenarios, which are common in VISOR. To address this, we adopt Cutie~\cite{cheng2024putting}, a long-term segmentation framework, as our baseline and replace its visual backbone with SAM2's pre-trained weights. This allows us to combine the long-term temporal modelling capability of the baseline with SAM2’s strong object perception and precise frame-to-frame correspondence learned from large-scale data. 

In summary, we combine fine-grained visual and geometric foundation models to tackle the unique challenges of egocentric video object segmentation. This design highlights the benefit of jointly leveraging visual and depth representations. On the VISOR test set, our method achieves a strong $\mathcal{J}\&\mathcal{F}$ score of 90.1\%.

\section{Related Works}
\label{sec:relate}

\textbf{Semi-Supervised Video Object Segmentation} (SVOS) aims to segment objects of interest in videos, where the targets are indicated by the annotations of the first frame by humans~\cite{gao2023deep}. Early works relied on propagation of labels through online fine-tuning~\cite{caelles2017one} or matching strategies~\cite{voigtlaender2019feelvos}, to transfer the initial mask and previous predictions to the current frame. The introduction of memory-based methods significantly improved segmentation performance by enabling a more effective use of intermediate predictions. These methods selectively store features and masks from intermediate frames (between the first and current frames) as memory, and retrieve relevant information to guide segmentation in the current frame. Following the initial memory-based models~\cite{oh2019video}, subsequent works have focused on improving memory management~\cite{liang2020video}, frame-wise affinity~\cite{cheng2021rethinking}, long-term memory~\cite{cheng2022xmem}, and object-aware memory design~\cite{cheng2024putting}. 

% % ------------------------------------
% ------------------------------------
\begin{figure*}[!ht]
\centering
\includegraphics[width=.8\linewidth]{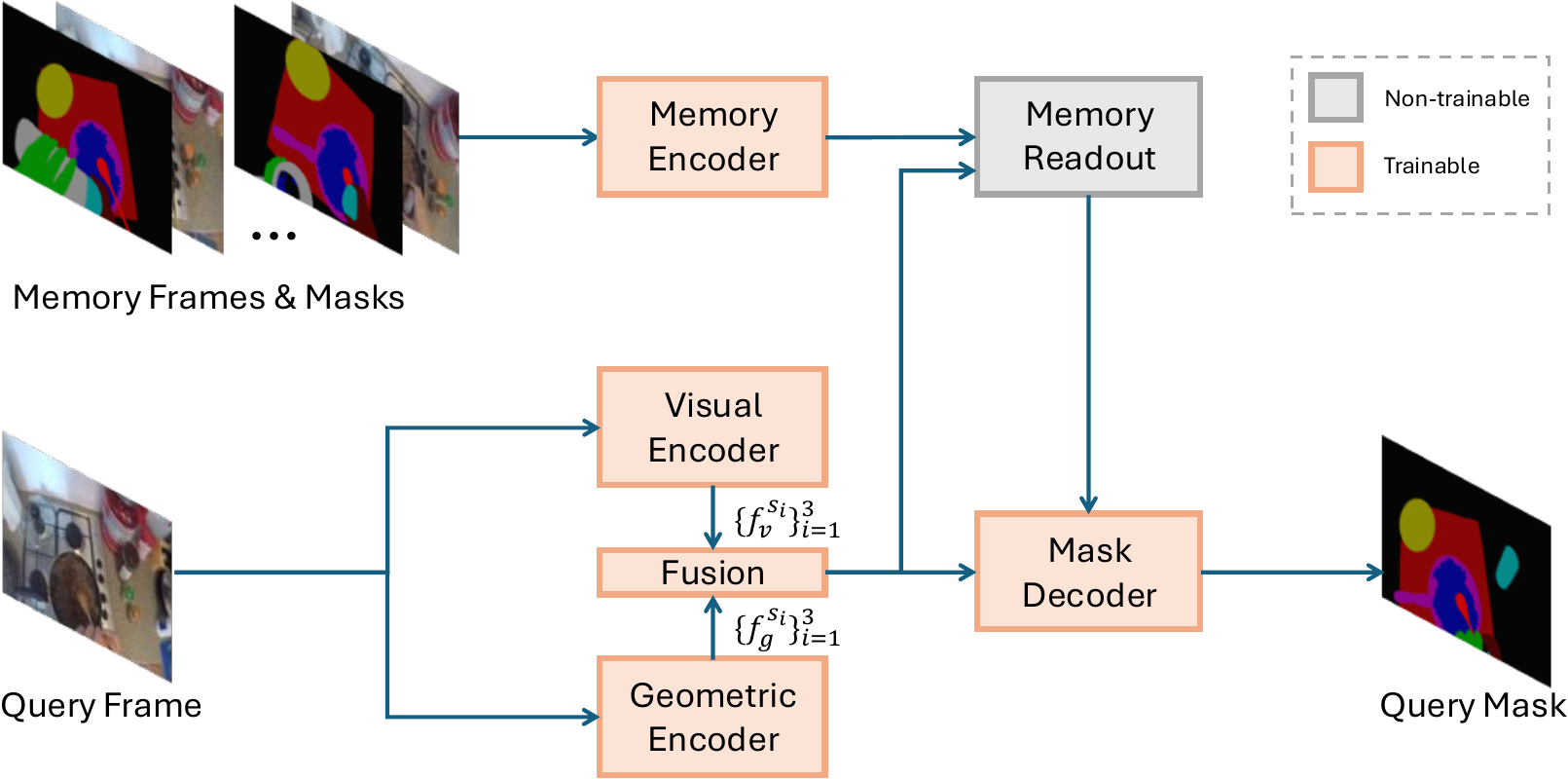}
   \caption{Overview of our solution. }
\label{fig:diagram}
\end{figure*}
% ------------------------------------

The success of SAM2~\cite{ravi2024sam} highlights that large-scale training can significantly improve segmentation performance, even with simple memory strategies that consider only the first and most recent frames. However, such limited memory usage is less effective for long-term videos. As a result, many follow-up works~\cite{videnovic2024distractor,yang2024samurai,ding2024sam2long} have explored better memory selection and utilisation to boost performance further. Although these approaches improve the results of existing benchmarks, most remain focused on visual cues, with limited attention to geometric information despite its importance in complex, dynamic scenes.

\section{Method}

Fig.~\ref{fig:diagram} shows our solution, where we consider Cutie~\cite{cheng2024putting} as the baseline, and we refer readers to the original paper for a comprehensive description. On top of the baseline, we consider Hiera-Large~\cite{ryali2023hiera}, with parameters pre-trained in SAM2~\cite{ravi2024sam}, as the visual encoder to enhance the object perception and frame-wise correspondence. In addition, we incorporate extra geometric cues to improve robustness against challenges in egocentric VOS. Specifically, the geometric encoder (DINOv2-Large~\cite{oquab2023dinov2} and a DPT decoder~\cite{ranftl2021vision} for multi-scale embeddings) comes from Depth Anything V2~\cite{depth_anything_v2}. 

Given a query frame $F_\mathrm{query} \in \mathbb{R}^{3 \times H \times W}$ of height $H$ and width $W$, the visual and geometric encoders extract multi-scale features, denoted as $\{f_\mathrm{v}^{s_i}\}_{i=1}^{3}$ and $\{f_\mathrm{g}^{s_i}\}_{i=1}^{3}$, where $s_1$, $s_2$, and $s_3$ correspond to 1/4, 1/8, and 1/16 of the original resolution, respectively.
Note that the patch sizes of DINOv2 and Hiera differ, which leads to mismatched feature resolutions between $f_\mathrm{v}$ and $f_\mathrm{g}$. To address this issue, we resize the input frames fed into the geometric encoder, ensuring that the resulting feature maps have consistent resolutions across different patch sizes. After feature encoding, $f_\mathrm{v}$ and $f_\mathrm{g}$ are concatenated at matching scales and subsequently fused via a learnable MLP layer. The fused features could be seamlessly integrated into the baseline framework and generate a mask for the query frame (we keep other modules the same as the baseline). 

\section{Experiments}

\subsection{Main Results}

Our training consists of two stages: 1) We freeze all encoders to initialise the fusion module and decoder on traditional VOS datasets, where we follow Cutie~\cite{cheng2024putting} to set data collections (the ``Mega'' version) and training coefficients. 2) To evaluate on the VISOR test set, we train our solution on the VISOR train and val sets for 100,000 iterations, with a batch size of 8, learning rate of 5e-5, and weight decay of 0.5. Due to sparse annotations and the long-term context in VISOR, we set the max skip as 1 when sampling frames as pseudo-training videos. 

Table~\ref{tab:test} shows our scores on the test set, where ``Hiera-L (SAM2)'' denotes the Hiera-Large and its initial parameters come from SAM2. ``MS'' indicates that we perform VOS under three different input frame sizes (1.2$\times$, 1.3$\times$, and 1.4$\times$), and ``Flip'' flags the use of flipped frames. In these cases of multiple inference, we sum all probabilities and generate the final predictions. 

\begin{table}[t]\renewcommand{\arraystretch}{1}
\small
\tabcolsep=0.15cm
  \centering
  \begin{tabular}{p{0.255\columnwidth}|p{0.09\columnwidth}<{\centering}|p{0.09\columnwidth}<{\centering}|p{0.112\columnwidth}<{\centering}p{0.112\columnwidth}<{\centering}p{0.112\columnwidth}<{\centering}}
  \toprule
  Backbone & Flip & MS & $\mathcal{J}\&\mathcal{F}$ & $\mathcal{J}$ & $\mathcal{F}$\\
  \noalign{\vspace{1.5pt}}
    \hline
    \noalign{\vspace{1.5pt}}
    Hiera-L (SAM2) & \ding{51} & \textcolor{gray!50}{\ding{55}} & 89.7\% & 87.5\% & 91.8\%\\
    Hiera-L (SAM2) & \ding{51} & \ding{51} & \cellcolor{blue!17!white} 90.1\% & 88.1\%\cellcolor{blue!17!white}  & 92.0\%\cellcolor{blue!17!white} \\
  \bottomrule
  \end{tabular}
  \caption{Ablations on the VISOR test set.}
  \label{tab:test}
\end{table}

\subsection{Ablations}

This section investigates the impact of different components of our solution on the segmentation performance. For simplicity and efficiency, all ablation variants are trained for 50,000 iterations with a batch size of 4, while keeping all other hyperparameters identical to those used in the main training setup. All results are shown in Table~\ref{tab:ablations}.

\paragraph{Impact of SAM2 Pretraining.} 
We first examine the effect of initialising the visual backbone with SAM2. This change alone leads to a substantial performance boost, which we attribute to the strong object perception and inter-frame correspondence priors learned from large-scale training. In addition, we fine-tune the original SAM2 model directly on VISOR. While it achieves competitive results ($\mathcal{J}\&\mathcal{F}$: 87.8\%), its performance remains below that of our full method w/o ``Depth'' and ``Post'', indicating that effective memory design for complex backgrounds and long-term dynamics is still crucial.

\paragraph{Benefit of Depth Integration.}
Next, we evaluate the contribution of the depth information. The results show performance improvements, confirming that depth provides valuable complementary cues.

\paragraph{Multi-Scale and Flip Fusion.}
Finally, we apply multi-scale inference and horizontal flip fusion during testing, which brings additional performance gains by improving prediction consistency.

\begin{table}[t]\renewcommand{\arraystretch}{1}
\small
\tabcolsep=0.15cm
  \centering
  \begin{tabular}{p{0.255\columnwidth}|p{0.09\columnwidth}<{\centering}|p{0.09\columnwidth}<{\centering}|p{0.112\columnwidth}<{\centering}p{0.112\columnwidth}<{\centering}p{0.112\columnwidth}<{\centering}}
  \toprule
  Backbone & Depth & Post & $\mathcal{J}\&\mathcal{F}$ & $\mathcal{J}$ & $\mathcal{F}$\\
  \noalign{\vspace{1.5pt}}
    \hline
    \noalign{\vspace{1.5pt}}
    Hiera-L (MAE) & \textcolor{gray!50}{\ding{55}} & \textcolor{gray!50}{\ding{55}} & 86.9\% & 85.6\% & 88.2\%\\
    Hiera-L (SAM2) & \textcolor{gray!50}{\ding{55}} & \textcolor{gray!50}{\ding{55}} & 88.3\% & 86.6\% & 89.9\%\\
    Hiera-L (SAM2) & \ding{51} & \textcolor{gray!50}{\ding{55}} & 88.7\% & 86.9\% & 90.5\% \\
    Hiera-L (SAM2) & \ding{51} & \ding{51} & \cellcolor{blue!17!white} 88.9\% & \cellcolor{blue!17!white} 87.0\% & \cellcolor{blue!17!white} 90.8\% \\
  \bottomrule
  \end{tabular}
  \caption{Ablations on the VISOR val set. The \colorbox{blue!17!white}{highlighted scores}  and the bottom results in Table~\ref{tab:test} come from the same setting. }
  \label{tab:ablations}
\end{table}

\section{Conclusion}

In this report, we address the challenges of egocentric VOS by leveraging large-scale visual pre-training and geometric depth cues. Through extensive experiments and ablation studies, we demonstrate that combining high-quality training priors from SAM2 with depth-aware design significantly improves segmentation accuracy, especially in complex and long-term scenarios. Our results highlight the importance of data scale and spatial understanding in advancing egocentric video understanding.

{
    \small
    \bibliographystyle{ieeenat_fullname}
    \bibliography{main}
}


\end{document}